\documentclass[10pt,twocolumn,letterpaper]{article}
\usepackage{iccv}
\usepackage{times}
\usepackage{epsfig}
\usepackage{graphicx}
\usepackage{amsmath}
\usepackage{amssymb}
\usepackage{multirow}

\DeclareMathOperator*{\argmax}{arg\,max}

\usepackage[pagebackref=true,breaklinks=true,colorlinks,bookmarks=false]{hyperref}

\iccvfinalcopy 

\def\iccvPaperID{2434} 

\begin{document}

\title{Online Multi-Object Tracking Using CNN-based Single Object Tracker with Spatial-Temporal Attention Mechanism}

\author{Qi Chu$^{1,3}$, Wanli Ouyang$^{2,3}$, Hongsheng Li$^3$, Xiaogang Wang$^3$, Bin Liu$^1$, Nenghai Yu$^{1,}$\thanks{Corresponding author.}\\
$^1$University of Science and Technology of China, $^2$University of Sydney\\
$^3$Department of Electronic Engineering, The Chinese University of Hong Kong\\
{\tt\small kuki@mail.ustc.edu.cn, \{wlouyang,hsli,xgwang\}@ee.cuhk.edu.hk, \{flowice,ynh\}@ustc.edu.cn}
}

\maketitle
\thispagestyle{empty}

\begin{abstract}
In this paper, we propose a CNN-based framework for online MOT. This framework utilizes the merits of single object trackers in adapting appearance models and searching for target in the next frame. Simply applying single object tracker for MOT will encounter the problem in computational efficiency and drifted results caused by occlusion. Our framework achieves computational efficiency by sharing features and using ROI-Pooling to obtain individual features for each target. Some online learned target-specific CNN layers are used for adapting the appearance model for each target. In the framework, we introduce spatial-temporal attention mechanism (STAM) to handle the drift caused by occlusion and interaction among targets.
The visibility map of the target is learned and used for inferring the spatial attention map. The spatial attention map is then applied to weight the features. Besides, the occlusion status can be estimated from the visibility map, which controls the online updating process via weighted loss on training samples with different occlusion statuses in different frames. It can be considered as temporal attention mechanism.
The proposed algorithm achieves 34.3\% and 46.0\% in MOTA on challenging MOT15 and MOT16 benchmark dataset respectively.
\end{abstract}

\section{Introduction}
\label{sec:intro}
Tracking objects in videos is an important problem in computer vision which has attracted great attention. It has various applications such as video surveillance, human computer interface and autonomous driving. The goal of multi-object tracking (MOT) is to estimate the locations of multiple objects in the video and maintain their identities consistently in order to yield their individual trajectories. MOT is still a challenging problem, especially in crowded scenes with frequent occlusion, interaction among targets and so on.

On the other hand, significant improvement has been achieved on single object tracking problem, sometimes called ``visual tracking'' in previous work. Most state-of-the-art single object tracking methods aim to online learn a strong discriminative appearance model and use it to find the location of the target within a search area in next frame \cite{MIL,struck,KCF,SRDCF}. Since deep convolutional neural networks (CNNs) are shown to be effective in many computer vision applications \cite{krizhevsky2012imagenet,fastrcnn,deepid,ouyang2015learning,zhang2015cross,yi2015understanding}, many works \cite{wang2015visual,hong2015online,ma2015hierarchical,STCT} have explored the usage of CNNs to learn strong discriminative appearance model in single object tracking and demonstrated state-of-the-art performance recently. An intuitive thought is that applying the CNN based single object tracker to MOT will make sense.

\begin{figure}[t]
  \begin{center}
    \includegraphics[width=1.0\linewidth]{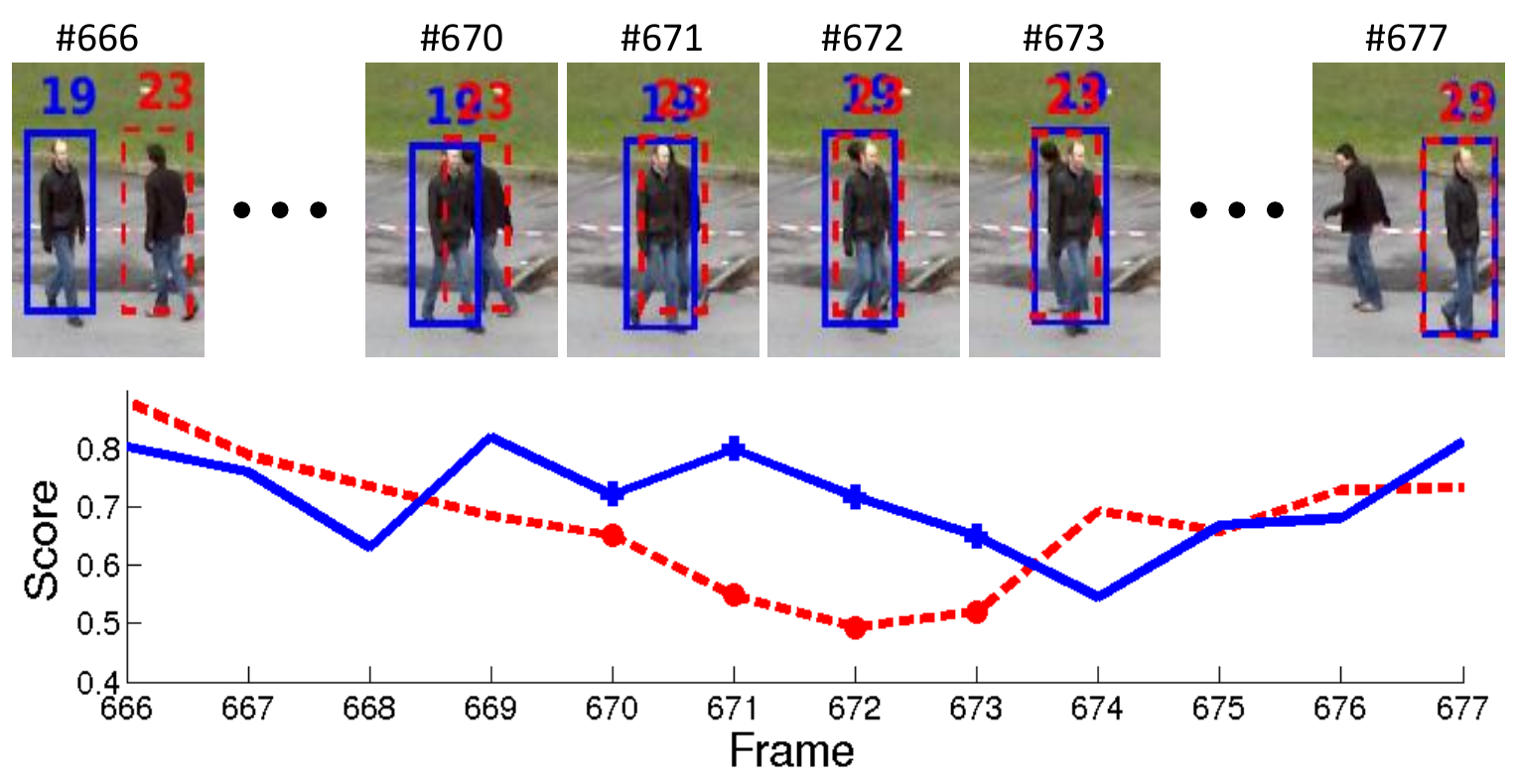}
  \end{center}
  \caption{An example of drift caused by occlusion of other targets when directly adopting single object trackers to MOT.}
  \label{fig:fig1}
\end{figure}

However, problems are observed when directly using single object tracking approach for MOT.

First, single object tracker may learn from noisy samples. In single object tracking,  the training samples for learning appearance model are collected online, where labels are based on tracking results. The appearance model is then used for finding the target in the next frame. When the target is occluded, the visual cue is unreliable for learning the appearance model. Consequently, the single object tracker will gradually drift and eventually fail to track the target. This issue becomes even more severe in MOT due to more frequent occlusion caused by interaction among targets. An example is shown in Figure \ref{fig:fig1}, one target is occluded by another when they are close to each other, which makes the visual cues of the occluded target contaminated when this target is used for training. However, the tracking score of the occluded target is still relatively high at the beginning of occlusion. In this case, the corresponding single object tracker updates the appearance model with the corrupted samples and gradually drifts to the occluder.

Second, since a new single object tracker needs to be added into MOT system once a new target appears, the computational cost of applying single object trackers to MOT may grow intolerably as the number of tracked objects increases, which limits the application of computationally intensive single object trackers in MOT such as deep learning based methods.


In this work, we focus on handling the problems observed above. To this end, we propose a dynamic CNN-based framework with spatial-temporal attention mechanism (STAM) for online MOT. In our framework, each object has its own individual tracker learned online. 

The contributions of this paper are as follows:

First, an efficient CNN-based online MOT framework. It solves the problem in computational complexity when simply applying CNN based single object tracker for MOT by sharing computation among multiple objects.

Second, in order to deal with the drift caused by occlusion and interactions among targets, spatial-temporal attention of the target is learned online. In our design, the visibility map of the target is learned and used for inferring the spatial attention map. The spatial attention map is applied to weight the features. Besides, the visibility map also indicates occlusion status of the target which is an important cue that needs to be considered in online updating process. The more severe a target is occluded, the less likely it should be used for updating corresponding individual tracker. It can be considered as temporal attention mechanism. Both the spatial and temporal attention mechanism help to help the tracker to be more robust to drift.

We demonstrate the effectiveness of the proposed online MOT algorithm, referred as STAM, using challenging MOT15 \cite{MOTchallenge} and MOT16 \cite{MOT16} benchmarks. 

\begin{figure*}[t]
  \begin{center}
    \includegraphics[width=0.95\linewidth]{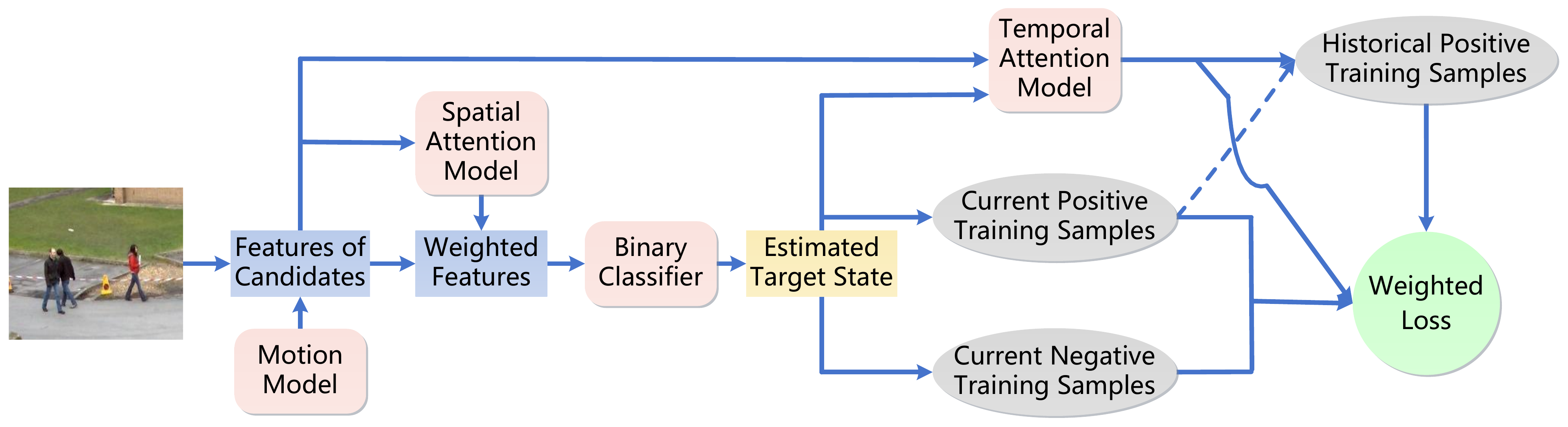}
  \end{center}
  \caption{Overview of the proposed algorithm STAM. Motion model provides the search area, where features of candidates are extracted and then weighted by the spatial attention. The candidate state with the maximum classification score is used as the estimated target state. The positive and negative training samples at current frame are collected according to the overlap with estimated target state.
  The historical positive training samples of the target are also used for online updating. Temporal attention model is used for weighting the loss of positive training samples in current and historical frames.}
  \label{fig:overview}
\end{figure*}

\section{Related Work}
{\bf Multi-object Tracking by Data Associtation.} With the development of object detection methods \cite{hog, DPM, fastrcnn, ouyang2016factors, ouyang2016learning}, data association \cite{hierarchical, DP_NMS, CEM, TC_ODAL} has become popular for MOT. The main idea is that a pre-defined object detector is applied to each frame, and then trajectories of objects are obtained by associating object detection results. Most of these works adopt an off-line way to process video sequences in which the future frames are also utilized to deal with the problem. These off-line methods consider MOT as a global optimization problem and focus on designing various optimization algorithm such as network flow \cite{DP_NMS,zhang2008global}, continuous energy minimization \cite{CEM}, max weight independent set \cite{MWIS}, k-partite graph \cite{GMCP,GMMCP}, subgraph multi-cut \cite{tang2015subgraph,JMC} and so on. However, offline methods are not suitable for causal applications such as autonomous driving. On the contrary, online methods generate trajectories only using information up to the current frame which adopt probabilistic inference \cite{oh2009markov} or deterministic optimization (\eg Hungarian algorithm used in \cite{TC_ODAL}). One problem of such association based tracking methods is the heavy dependency on the performance of the pre-defined object detector. This problem has more influence for online tracking methods, since they are more sensitive to noisy detections. Our work focuses on applying online single object tracking methods to MOT. The target is tracked by searching for the best matched location using online learned appearance model. This helps to alleviate the limitations from imperfect detections, especially for missing detections. It is complementary to data association methods, since the tracking results of single object trackers at current frame can be consider as association candidates for data association.

{\bf Single Object Tracker in MOT.} Some previous works \cite{xing2009multi,yang2012online,breitenstein2011online,yan2012track,zhang2013structure,MDP} have attempted to adopt single object tracking methods into MOT problem. However, single object tracking methods are often used to tackle a small sub-problem due to challenges mentioned in Sec. \ref{sec:intro}. For example, single object trackers are only used to generate initial tracklets in \cite{xing2009multi}.
Yu \etal \cite{MDP} partitions the state space of the target into four subspaces and only utilizes single object trackers to track targets in tracked state. There also exists a few works that utilize single object trackers throughout the whole tracking process. Breitenstein \etal \cite{breitenstein2011online} use target-specific classifiers to compute the similarity for data association in a particle filtering framework. Yan \etal \cite{yan2012track} keep both the tracking results of single object trackers and the object detections as association candidates and select the optimal candidate using an ensemble framework. All methods mentioned above do not make use of CNN based single object trackers, so they can not update features during tracking. Besides, they do not deal with tracking drift caused by occlusion. Different from these methods, our work adopts online learned CNN based single object trackers into online multi-object tracking and focuses on handling drift caused by occlusion and interactions among targets.


{\bf Occlusion handling in MOT.} Occlusion is a well-known problem in MOT and many approaches are proposed for handling occlusion. Most works \cite{hu2012single,wu2007detection,shu2012part,izadinia20122t,tang2014detection} aim at utilizing better detectors for handling partial occlusion. In this work, we attempt to handle occlusion from the perspective of feature learning, which is complementary to these detection methods. Specifically, we focus on learning more robust appearance model for each target using the single object tracker with the help of spatial and temporal attention.

\section{Online MOT Algorithm}
\subsection{Overview}
The overview of the proposed algorithm is shown in Figure \ref{fig:overview}. The following steps are used for tracking objects:

Step 1. At the current frame $t$, the search area of each target is obtained using motion model. The candidates are sampled within the search area.

Step 2. The features of candidates for each target are extracted using ROI-Pooling and weighted by spatial attention. Then the binary classifier is used to find the best matched candidate with the maximum score, which is used as the estimated target state.

Step 3. The visibility map of each tracked target is inferred from the feature of corresponding estimated target state. The visibility map of the tracked target is then used along with the spatial configurations of the target and its neighboring targets to infer temporal attention.

Step 4. The target-specific CNN branch of each target is updated according to the loss of training samples in current and historical frames weighted by temporal attention. The motion model of each target is updated according to corresponding estimated target state.

Step 5. The object management strategy determines the initialization of new targets and the termination of untracked targets.

Step 6. If frame $t$ is not the last frame, then go to Step 1 for the next frame $t+1$ .

\begin{figure*}[t]
  \begin{center}
    \includegraphics[width=0.95\linewidth]{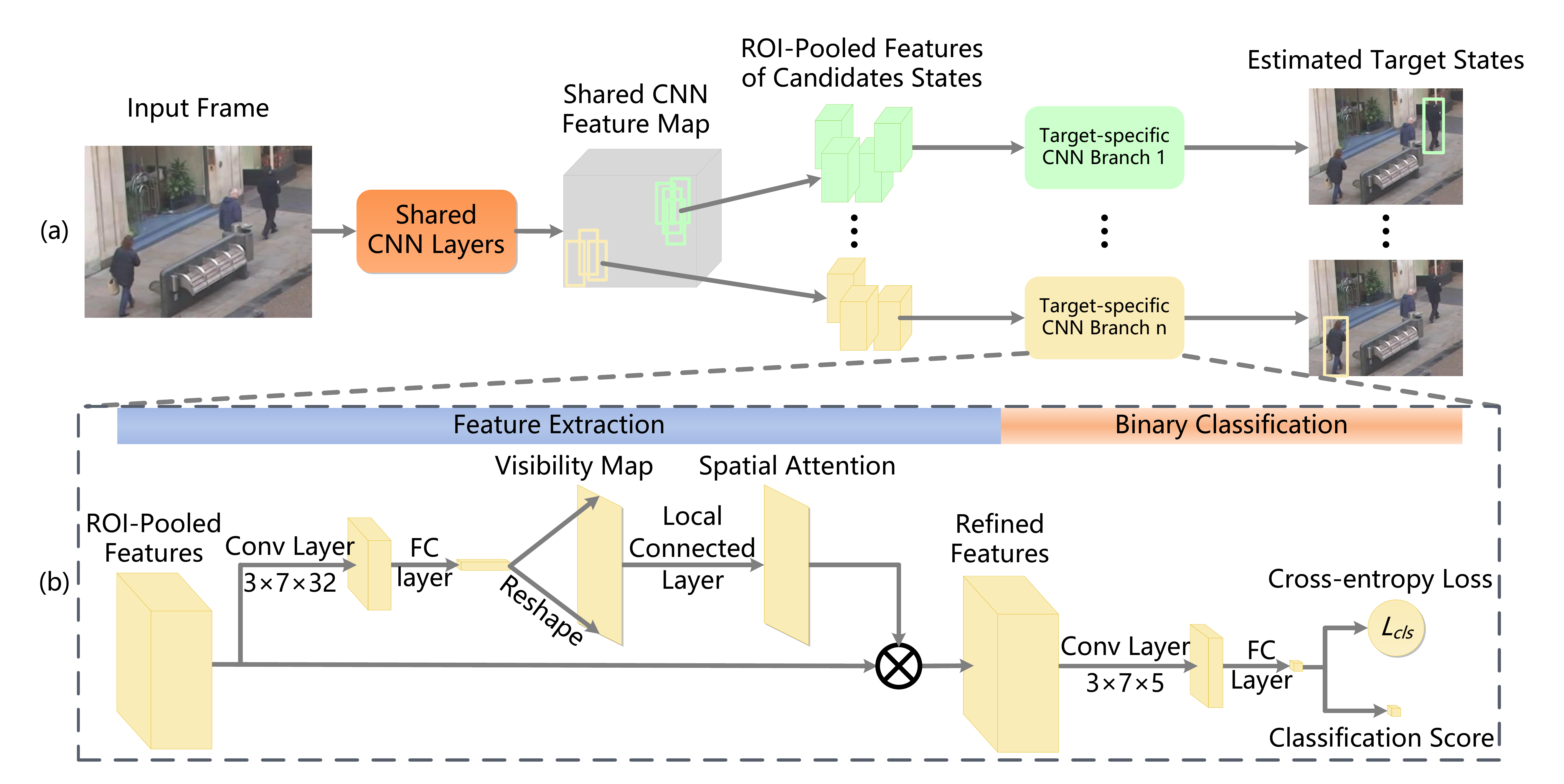}
  \end{center}
  \caption{(a) The framework of the proposed CNN model. It contains shared CNN layers and multiple target-specific CNN branches. The shared layers are shared by all targets to be tracked. Each target has its own corresponding target-specific CNN branch which is learned online. The target-specific CNN branch acts as a single object tracker and can be added to or removed from the whole model according to the entrance of new target or exit of existing target. (b) The details of the target-specific CNN branch. Each target-specific CNN branch consists of feature extraction using visibility map and spatial attention as described in Sec. \ref{subsec:featureExtraction} and binary classification (described in Sec. \ref{Sec:ClsDet}). The initialization and online updating of the target-specific branch are described in Sec. \ref{sec:initialization} and Sec. \ref{sec:update_appearance} respectively.}
  \label{fig:framework}
\end{figure*}

\subsection{Dynamic CNN-based MOT Framework}


We propose a dynamic CNN-based framework for online MOT, which consists of both shared CNN layers and target-specific CNN branches. As shown in Figure \ref{fig:framework}, the shared CNN layers encode the whole input frame as a large feature map, from which the feature representation of each target is extracted using ROI-Pooling \cite{fastrcnn}. For computational efficiency, these shared layers are pre-trained on Imagenet Classification task \cite{imagenet}, and not updated during tracking. All target-specific CNN branches share the same structure, but are separately trained to capture the appearance of different targets. They can be viewed as a set of single-object trackers.


The number of target-specific CNN branches varies with the number of existing targets. Once a new target appears, a new branch will be initialized and added to the model.
If a target is considered to be disappeared, its corresponding branch will be removed from the entire model.

\subsection{Online Tracking with STAM}
\label{Sec:Tracking}
The trajectory of an object can be represented by a series of states denoted by $\{ \mathbf{x}_t \}_{ t=1,2,3...,T}$, where $\mathbf{x}_t=[x_t,y_t,w_t,h_t]$. $x_t$ and $y_t$ represent the center location of the target at frame $t$. $w_t$ and $h_t$ denote the width and height of the target, respectively. 
Multi-object tracking aims to obtain the estimated states of all targets at each frame.

\subsubsection{Candidate States}
For the $i$-th target $\mathcal{T}^i$ to be tracked, its estimated state $ \mathbf{x}_{t}^i$ at frame $t$ is obtained by searching from a set of candidate states denoted by $\mathcal{C}_{t}^i$, which consists of two subsets:
\begin{equation}
 \mathcal{C}_{t}^i= \{\mathbf{x}_{t,n}^s\}^{N_i}_{n=1} \bigcup \mathcal{D}_{t}^i,
\end{equation}
$\{\mathbf{x}_{t,n}^s\}^{N_i}_{n=1}$ denotes the set of candidate states that are drawn from a Gaussian distribution $\mathcal{N}(\tilde{\mathbf{x}}_{t}^i, \Sigma_t^i)$, where $\tilde{\mathbf{x}}_{t}^i$ is the predicted state of target $\mathcal{T}^i$ at frame $t$, and $\Sigma_t^i=diag\left((\sigma_{t,x}^i)^2,(\sigma_{t,y}^i)^2,(\sigma_{t,w}^i)^2,(\sigma_{t,h}^i)^2 \right)$ is a diagonal covariance matrix indicating the variance of target location and scale. $\tilde{\mathbf{x}}_{t}^i$ and $\Sigma_t^i$  are estimated by the motion model (Sec. \ref{sec:update_motion}).
Denote by $\mathcal{D}_t=\{\mathbf{x}_{t,m}^d\}_{m=1}^M$ the set of all object detections provided by an off-line trained detector at frame $t$. $\mathcal{D}_{t}^i=\{\mathbf{x}_{t,m_i}^d\}^{M_i}_{m_i=1}\subseteq\mathcal{D}_t$ are selected detections that are close to the predicted state $\tilde{\mathbf{x}}_{t}^i$ in spatial location ($|(\mathbf{x}_{t,m_i}^d)_k-(\tilde{\mathbf{x}}_{t}^i)_k|<3\sigma_{t,k}^i, \forall k=x,y,w,h$).

\subsubsection{Feature Extraction with Spatial Attention}
\label{subsec:featureExtraction}
The feature of candidate state is extracted from the shared feature map using ROI-Pooling and spatial attention mechanism. The ROI-Pooling from the shared feature map ignores the fact that the tracked targets could be occluded. In this case, the pooled features would be distorted by the occluded parts. To handle this problem, we propose a spatial attention mechanism which pays more attention to un-occluded regions for feature extraction.

Directly using spatial attention does not work well due to limited training samples in the online learning process. In our work, we first generate the visibility map which encodes the spatial visibility of the input samples. Then the spatial attention is derived from visibility map.

\begin{figure}[t]
  \begin{center}
    \includegraphics[width=1.0\linewidth]{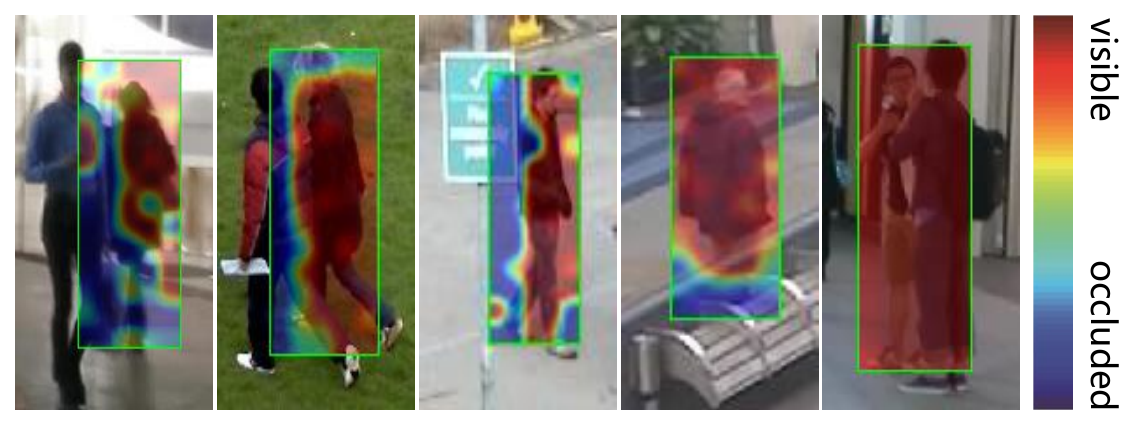}
  \end{center}
  \caption{Examples of the generated visibility maps. The first four columns show examples of the target occluded by other target or background. The last column shows the failure case when targets are too close. Best viewed in color.}
  \label{fig:vis}
\end{figure}

{\bf Visibility Map.} Denote the ROI-Pooled feature representation of the $j$-th candidate state $\mathbf{x}_{t,j}^i\in\mathcal{C}_t^i$ as $\Phi_{roi}(\mathbf{x}_{t,j}^i)\in \mathbb{R}^{W\times H\times C}$, the visibility map of $\mathbf{x}_{t,j}^i$ is estimated as
\begin{equation}
\label{eq:visibility}
\mathbf{V}(\mathbf{x}_t^j)=f_{vis}(\Phi_{roi}(\mathbf{x}_t^j);\boldsymbol{w}_{vis}^i),~~~\mathbf{V}(\mathbf{x}_t^j)\in \mathbb{R}^{W\times H}
\end{equation}
where, $\boldsymbol{w}_{vis}^i$ is the set of parameters. $f_{vis}(\Phi_{roi}(\mathbf{x}_{t,j}^i);\boldsymbol{w}_{vis}^i)$ is modeled as two layers interleaved with ReLU layer. The first layer is a convolution layer which has the kernel size of $3 \times 7$ and produces a feature map with 32 channels. The second layer is a fully connected layer with the output size of $(W*H)$. Then the output is reshaped to a map with the size of $W\times H$. Each element in visibility map $\mathbf{V}(\mathbf{x}_{t,j}^i)$ indicates the visibility of corresponding location in feature map $\Phi_{roi}(\mathbf{x}_{t,j}^i)$. Some examples of generated visibility maps are shown in Figure \ref{fig:vis}.


{\bf Spatial Attention.} The spatial attention map $\Psi(\mathbf{x}_{t,j}^i)\in \mathbb{R}^{W\times H}$ for candidate state $\mathbf{x}_{t,j}^i$ is obtained from visibility map $\mathbf{V}(\mathbf{x}_{t,j}^i)$ as follows:
\begin{equation}
\label{eq:spatial_attenton_1}
\Psi(\mathbf{x}_{t,j}^i)) =f_{att}(\mathbf{V}(\mathbf{x}_{t,j}^i); \boldsymbol{w}_{att}^i), \\
\end{equation}
where $f_{att}$ is implemented by a local connected layer followed by a spatial softmax layer and  $\boldsymbol{w}_{att}^i$ denotes the parameters. Then the spatial attention map $\Psi(\mathbf{x}_{t,j}^i)$ is applied to weight the feature map $\Phi_{roi}(\mathbf{x}_{t,j}^i)$ as
\begin{equation}
\label{eq:spatial_attenton}
\begin{aligned}
\Phi_{att}(\mathbf{x}_{t,j}^i)& =\Phi_{roi}(\mathbf{x}_{t,j}^i) \star \Psi(\mathbf{x}_{t,j}^i), \\
\Phi_{att}(\mathbf{x}_{t,j}^i)&, \Phi_{roi}(\mathbf{x}_{t,j}^i)\in \mathbb{R}^{W\times H\times C} \\
\Psi(\mathbf{x}_{t,j}^i)&\in \mathbb{R}^{W\times H}
\end{aligned}
\end{equation}
where $\star$ represents the channel-wise Hadamard product operation, which performs Hadamard product between $\Psi(\mathbf{x}_{t,j}^i)$ and each channel of $\Phi_{roi}(\mathbf{x}_{t,j}^i)$.

\subsubsection{Target State Estimation Using Binary Classifier and Detection Results}
\label{Sec:ClsDet}
\textbf{Binary Classification.} Given the refined feature representation $\Phi_{att}(\mathbf{x}_{t,j}^i)$, the classification score is obtained as follows:
\begin{equation}
  \label{eq:loc2}
  p_{t,j}^i= f_{cls}(\Phi_{att}(\mathbf{x}_{t,j}^i);\boldsymbol{w}_{cls}^i),
\end{equation}
where $p_{t,j}^i \in [0,1]$ is the output of binary classifier which indicates the probability of candidate state $\mathbf{x}_{t,j}^i$ belonging to target $\mathcal{T}^i$. $\boldsymbol{w}_{cls}^i$ is the parameter of the classifier for target $\mathcal{T}^i$. In our work, $f_{cls}(\Phi_{att}(\mathbf{x}_{t,j}^i);\boldsymbol{w}_{cls}^i)$ is modeled by two layers interleaved with ReLU layer. The first layer is a convolution layer which has the kernel size of $3 \times 7$ and produces a feature map with 5 channels. The second layer is a fully connected layer with the output size of 1. Then a sigmoid function is applied to ensure the output to be in $[0,1]$.

The primitive estimated state of target $\mathcal{T}^i$ is obtained by searching for the candidate state with the maximum classification score as follows:
\begin{equation}
  \label{eq:loc1}
  \hat{\mathbf{x}}^i_t=\argmax_{\mathbf{x}_{t,j}^i\in\mathcal{C}_t^i}f_{cls}(\Phi_{att}(\mathbf{x}_{t,j}^i);\boldsymbol{w}_{cls}^i),
\end{equation}

{\bf State Refinement.} The primitive estimated state with too low classification score will bias the updating of the model. To avoid model degeneration, if the score $\hat{y}_{t}^i=f_{cls}(\Phi_{att}(\hat{\mathbf{x}}^i_t;\boldsymbol{w}_{cls}^i)$ is lower than a threshold $p_0$, the corresponding target $\mathcal{T}^i$ is considered as ``untracked'' in current frame $t$. Otherwise, the primitive state $\hat{\mathbf{x}}^i_t$ will be further refined using the object detections states $\mathcal{D}_t=\{\mathbf{x}_{t,m}^d\}_{m=1}^M$.

Specifically, the nearest detection state  for $\hat{\mathbf{x}}^i_t$ is obtained as follows:
\begin{equation}
\mathbf{x}_{t}^{d,i}=\argmax_{\mathbf{x}_{t,m}^d \in \mathcal{D}_t}\ IoU(\hat{\mathbf{x}}^i_t,\mathbf{x}_{t,m}^d),
\end{equation}
where $IoU(\hat{\mathbf{x}}^i_t,\mathbf{x}_{t,m}^d)$ calculates the bounding box IoU overlap ratio between $\hat{\mathbf{x}}^i_t$ and $\mathbf{x}_{t,m}^d$. Then the final state of target $\mathcal{T}^i$ is refined as
\begin{equation}
  \label{eq:loc2}
  \begin{split}
    &\mathbf{x}^i_t=
    \begin{cases}
      o_t^i\ \mathbf{x}_{t}^{d,i}+(1-o_t^i)\hat{\mathbf{x}}^i_t, & o_t^i>o_0\\ 
      \qquad\qquad\hat{\mathbf{x}}^i_t, & otherwise,
    \end{cases}\\
  \end{split}
\end{equation}
where $o_t^i=IoU(\hat{\mathbf{x}}^i_t,\mathbf{x}_{t}^{d,i})$ and $o_0$ is a pre-defined threshold.

\subsection{Model Initialization and Online Updating}
Each target-specific CNN branch comprises of visibility map, attention map and binary classifier. The parameters for visibility map are initialized in the first frame when the target appears and then all three modules are jointly learned.
\vspace{-10pt}
\subsubsection{Model Initialization}
\label{sec:initialization}
For the initialization of parameters in obtaining visibility map, we synthetically generate training samples and the corresponding ground truth based on initial target state.

{\bf Augmented Set.} Denote the ROI-Pooled feature representation of initial state of target $\mathcal{T}^{i}$ as $\Phi_{roi}(\mathbf{x}_0^{i}) \in \mathbb{R}^{W\times H\times C}$, a $W\times H$ matrix with all elements equal to $1$ is used as the corresponding ground truth visibility map. An augmented set is obtained via collecting samples that have large overlap with initial target state $\mathbf{x}_0^{i}$. For each sample in the augmented set, the ground truth visibility map for region not overlapping with $\mathbf{x}_0^{i}$ is set to $0$.

{\bf Feature Replacement.} We replace the features of the sample with the features from another target or background at some region and set the ground truth for replaced region to $0$. The replaced region is regarded as occluded. For each sample in the augmented set, the feature replacement is done using different targets/brackgrounds at different regions.

Given these training samples and ground truth visibility maps, the model is trained using cross-entropy loss.
\vspace{-5pt}
\subsubsection{Online Updating Appearance Model}
\label{sec:update_appearance}
After initialization in the initial frame, all three modules are jointly updated during tracking using back-propagation algorithm.

Training samples used for online updating are obtained from current frame and historical states. For tracked target, positive samples at current frame $t$ are sampled around the estimated target state $\mathbf{x}_t$ with small displacements and scale variations. Besides, historical states are also utilized as positive samples. If the target is considered as "untracked" at current frame, we only use historical states of the target as positive samples. All negative samples are collected at current frame $t$. The target-specific branch needs to have the capability of discriminating the target from other targets and background. So both the estimated states of other tracked targets and the samples randomly sampled from background are treated as the negative samples.

For target $\mathcal{T}^i$, given the current positive samples set $\{\mathbf{x}_{t,j}^{i+}\}_{j=1}^{N_t^{i+}}$, historical positive samples set $\{\mathbf{x}_{h,j}^{i+}\}_{j=1}^{N_{h}^{i+}}$ and the negative samples set $\{\mathbf{x}_{t,j}^{i-}\}_{j=1}^{N_{t}^{i-}}$, the loss function for updating corresponding target-specific branch is defined as
\begin{equation}
\label{eq:basic_loss}
\begin{aligned}
L^i_t=L^{i-}_{t}+(1-\alpha^i_t )L^{i+}_{t}+\alpha^i_t L^{i+}_{h},
\end{aligned}
\end{equation}
\begin{equation}
\begin{aligned}
&L^{i-}_{t}=-\frac{1}{N_{t}^{i-}}\sum^{N_{t}^{i-}}_{j=1}\log[1-f_{cls}(\Phi_{att}(\mathbf{x}_{t,j}^{i-});\boldsymbol{w}_{cls}^i)],\\
&L^{i+}_{t}=-\frac{1}{N_{t}^{i+}}\sum^{N_{t}^{i+}}_{j=1}\log f_{cls}(\Phi_{att}(\mathbf{x}_{t,j}^{i+});\boldsymbol{w}_{cls}^i),\\
&L^{i+}_{h}=-\frac{1}{N_{h}^{i+}}\sum^{N_{h}^{i+}}_{j=1}\log f_{cls}(\Phi_{att}(\mathbf{x}_{h,j}^{i+});\boldsymbol{w}_{cls}^i),
\end{aligned}
\end{equation}
where, $L^{i-}_{t}$, $L^{i+}_{t}$, and $L^{i+}_{h}$ are losses from negative samples, positive samples at current frame, and positive samples in the history, respectively. $\alpha^i_t$ is the temporal attention introduced below.

{\bf Temporal Attention.} A crucial problem for model updating is to balance the relative importance between current and historical visual cues. Historical samples are reliable positive samples collected in the past frames, while samples in current frame reflect appearance variations of the target. In this work, we propose a temporal attention mechanism, which dynamically pay attention to current and historical samples based on occlusion status.

Temporal attention of target $\mathcal{T}^i$ is inferred from visibility map $\mathbf{V}(\mathbf{x}_t^i)$ and the overlap statuses with other targets
\begin{equation}
\label{eq:occ_status}
\alpha^i_t=\sigma(\gamma^i s^i_t+\beta^i o^i_t+b^i),
\end{equation}
where $s^i_t$ is the mean value of visibility map $\mathbf{V}(\mathbf{x}_t^i)$. $o^i_t$ is the maximum overlap between $\mathcal{T}^i$ and all other targets in current frame $t$. $\gamma^i$, $\beta^i$ and $b^i$ are learnable parameters. $\sigma(x)=1/(1+e^{-x})$ is the sigmoid function.

Since $\alpha^i_t$ indicates the occlusion status of target $\mathcal{T}^i$. If $\alpha^i_t$ is large, it means that target $\mathcal{T}^i$ is undergoing severe occlusion at current frame $t$. Consequently, the weight for positive samples at current frame is small according to Eq. \ref{eq:basic_loss}. There, the temporal attention mechanism provides a good balance between current and historical visual cues of the target. Besides, if $\alpha^i_t$ is smaller than a threshold $\alpha_0$, the corresponding target state $\mathbf{x}_t^i$ will be added to the historical samples set of target $\mathcal{T}^i$.

\subsubsection{Updating Motion Model}
\label{sec:update_motion}
Most single object trackers do not consider the motion model, while it is proved to be helpful in MOT. In our work, a simple linear motion model with constant velocity and Gaussian noise is applied to each target, which is used to determine the center location and the size of search area for tracking the target in next frame. The scale of the target is considered as unchanged. Given the velocity $\boldsymbol{v}_t^i$ at frame $t$, the predicted state of target $\mathcal{T}^i$ at frame $t+1$ is defined as $\tilde{\mathbf{x}}_{t+1}^i=\mathbf{x}_{t}^i+[\boldsymbol{v}_t^i,0,0]$. 

At frame $t$, the velocity of target $\mathcal{T}^i$ is updated as
\begin{equation}
  \label{eq:v}
  \begin{split}
    &\tilde{\boldsymbol{v}}_t^i=\frac{1}{T_{gap}}(\boldsymbol{l}_t^i-\boldsymbol{l}_{t-T_{gap}}^i), \\
    &\boldsymbol{v}_t^i=\alpha_t^i\boldsymbol{v}_{t-1}^i+(1-\alpha_t^i)\tilde{\boldsymbol{v}}_t^i,
  \end{split}
\end{equation}
where $T_{gap}$ denotes the time gap for computing velocity. $\boldsymbol{l}_t^i=[x_t^i,y_t^i]^T$ is the center location of target $\mathcal{T}^i$ at frame $t$. The variance of Gaussian noise is defined as
\begin{equation}
  \label{eq:searchArea}
  \begin{split}
    &\sigma_{t,w}^i=\sigma_{t,h}^i=\tfrac{1}{30}h_t^i,\\
    &\sigma_{t,x}^i=\sigma_{t,y}^i=\sigma_t^i,\\
    &\sigma_t^i=
    \begin{cases}
        1.05\cdot\sigma_{t-1}^i, & \tilde{N}_t^i>0\\
        r\cdot\sigma_{t-1}^i/0.75, & \tilde{N}_t^i=0\ and\ r>0.75\\
        max(\tfrac{1}{20}h_t^i,\tfrac{1}{2}\sigma_{t-1}^i), & \tilde{N}_t^i=0\ and\ r<0.25\\
        \sigma_{t-1}^i, & otherwise
    \end{cases}\\
    &r=||\boldsymbol{l}_t^i-\tilde{\boldsymbol{l}}_t^i||_2/(3\sigma_{t-1}^i),
  \end{split}
\end{equation}
where $\tilde{\boldsymbol{l}}_t^i=\boldsymbol{l}_{t-1}^i+\boldsymbol{v}_{t-1}^i$ is the center location of target $\mathcal{T}^i$ at frame $t$ predicted by motion model. $\tilde{N}_t^i$ denotes the length of the successive untracked frames of target $\mathcal{T}^i$ at frame $t$. $r$ measures the prediction error of linear motion model. If target $\mathcal{T}^i$ is tracked at frame $t$, the variance $\sigma_t^i$ is related to the prediction error $r$. Otherwise, the search area will be extended as the length of successive untracked frames grows.


\subsection{Object Management}
\label{sec:management}
In our work, a new target $\mathcal{T}^{new}$ is initialized when a newly detected object with high detection score is not covered by any tracked targets. To alleviate the influence of false positive detections, the newly initialized target $\mathcal{T}^{new}$ will be discarded if it is considered as ``untracked'' (Sec. \ref{Sec:ClsDet}) or not detected in any of the first $T_{init}$ frames. For target termination, we simply terminate the target if it is ``untracked'' for over $T_{term}$ successive frames. Besides, targets that exit the field of view are also terminated.

\section{Experiments}
In this section, we present the experimental results and analysis for the proposed online MOT algorithm.

\subsection{Implementation details}
The proposed algorithm is implemented in MATLAB with Caffe \cite{caffe}. In our implementation, we use the first ten convolutional layers of the VGG-16 network \cite{VGG} trained on Imagenet Classification task \cite{imagenet} as the shared CNN layers. The threshold $o_0$ is set to 0.5, which determines whether the location found by single object tracker is covered by a object detection. The thresholds $p_0$ and $\alpha_0$ are set to 0.7 and 0.3 respectively. For online updating, we collect positive and negative samples with $\ge0.7$ and $\le0.3$ IoU overlap ratios with the target state at current frame, respectively. The detection scores are normalized to the range of $[0,1]$ and the detection score threshold in target initialization is set to 0.25. Denote the frame rate of the video as $F$, we use $T_{init}=0.2F$ and $T_{term}=2F$ in object management and $T_{gap}=0.3F$ in motion model.
%
\vspace{5pt}
\subsection{Datasets}
We evaluate our online MOT algorithm on the public available MOT15 \cite{MOTchallenge} and MOT16 \cite{MOT16} benchmarks containing 22 (11 training, 11 test) and 14 (7 training, 7 test) video sequences in unconstrained environments respectively. The ground truth annotations of the training sequences are released. We use the training sequences in MOT15 benchmark for performance analysis of the proposed method. The ground truth annotations of test sequences in both benchmarks are not released and the tracking results are automatically evaluated by the benchmark. So we use the test sequences in two benchmarks for comparison with various state-of-the-art MOT methods. In addition, these two benchmarks also provide object detections generated by the ACF detector \cite{ACF} and the DPM detector \cite{DPM} respectively. We use these public detections in all experiments for fair comparison.

\vspace{5pt}
\subsection{Evaluation metrics}
To evaluate the performance of multi-object tracking methods, we adopt the widely used CLEAR MOT metrics \cite{bernardin2008evaluating}, including multiple object tracking precision (MOTP) and multiple object tracking accuracy (MOTA) which combines false positives (FP), false negatives (FN) and the identity switches (IDS). Additionally, we also use the metrics defined in \cite{li2009learning}, which consists of the percentage of mostly tracked targets (MT, a ground truth trajectory that are covered by a tracking hypothesis for at least 80\% is regarded as mostly tracked), the percentage of mostly lost targets (ML, a ground truth trajectory that are covered by a tracking hypothesis for at most 20\% is regarded as mostly lost),
and the number of times a trajectory is fragmented (Frag).

\vspace{5pt}
\subsection{Tracking Speed}
The overall tracking speed of the proposed method on MOT15 test sequences is 0.5 fps using the 2.4GHz CPU and a TITAN X GPU, while the algorithm without feature sharing runs at 0.1 fps with the same environment.
\vspace{5pt}
\begin{figure}[t]
  \begin{center}
    \includegraphics[width=0.95\linewidth]{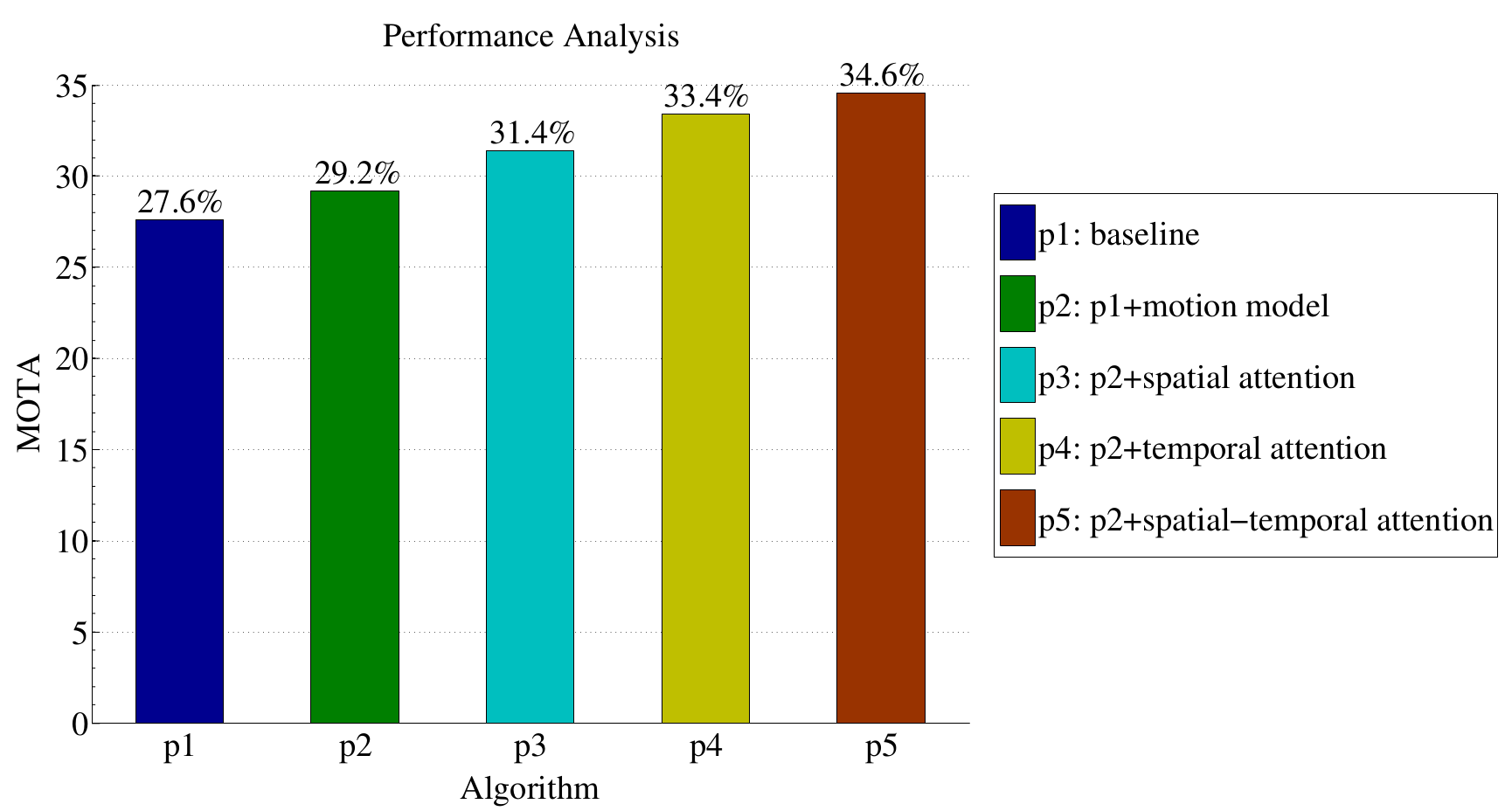}
  \end{center}
  \vspace{-5pt}
  \caption{The performance of different algorithms on training sequences of MOT15 in terms of MOTA.}
  \label{fig:MOTA}
\end{figure}
\vspace{-5pt}

\subsection{Performance analysis}
To demonstrate the effectiveness of the proposed method, we build five algorithms for components of different aspects of our approach. The details of each algorithm are described as follows:

p1: directly using single object trackers without the proposed spatial-temporal attention or motion model, which is the baseline algorithm;

p2: adding the motion model based on p1;

p3: adding the spatial attention based on p2;

p4: adding the temporal attention based on p2;

p5: adding the spatial-temporal attention based on p2, which is the whole algorithm with all proposed components.

The performance of these algorithms on the training sequences of MOT15, in terms of MOTA which is a good approximation of the overall performance, are shown in Figure \ref{fig:MOTA}. The better performance of the algorithm p2 compared to p1 shows the effect of the using motion model in MOT. The advantages of the proposed spatial-temporal attention can be seen by comparing the performance of algorithm p5 and p2. Furthermore, compared to the algorithm p2, the performance improvement of p3 and p4 shows the effectiveness of spatial and temporal attention in improving tracking accuracy respectively. The improvement of p5 over both p3 and p4 shows that the spatial and temporal attention are complementary to each other. Algorithm p5 with all the proposed components achieves the best performance and improves 8\% in terms of MOTA compared with the baseline algorithm p1, which demonstrates the effectiveness of our algorithm in handling the problems of using single object trackers directly.
\begin{table*}[t]
  \centering
  \begin{tabular}{|c|c|c|c|c|c|c|c|c|c|c|}
    \hline
    benchmark & Mode & Method & MOTA $\uparrow$ & MOTP $\uparrow$ & MT $\uparrow$ & ML $\downarrow$ & FP $\downarrow$ & FN $\downarrow$ & IDS $\downarrow$ & Frag $\downarrow$ \\
    \hline
    \hline
    \multirow{13}{*}{MOT15} & \multirow{7}{*}{Offline} & SMOT\cite{SMOT} & 18.2\% & 71.2\% & 2.8\% & 54.8\% & 8780 & 40310 & 1148 & 2132\\
    & & CEM\cite{CEM} & 19.3\% & 70.7\% & 8.5\% & 46.5\% & 14180 & 34591 & 813 & 1023\\
    & & JPDA\_m\cite{JPDA} & 23.8\% & 68.2\% & 5.0\% & 58.1\% & \textcolor{blue}{\bf{4533}} & 41873 & \textcolor{blue}{\bf{404}} & \textcolor{blue}{\bf{792}}\\
    & & SiameseCNN\cite{SiameseCNN} & 29.0\% & 71.2\% & 8.5\% & 48.4\% & 5160 & 37798 & 639 & 1316\\
    & & CNNTCM\cite{CNNTCM} & 29.6\% & 71.8\% & 11.2\% & 44.0\% & 7786 & 34733 & 712 & 943\\
    & & MHT\_DAM\cite{MHT} & 32.4\% & 71.8\% & \textcolor{blue}{\bf{16.0\%}} & \textcolor{blue}{\bf{43.8\%}} & 9064 & \textcolor{blue}{\bf{32060}} & 435 & 826\\
    & & NOMT\cite{NOMT} & \textcolor{blue}{\bf{33.7\%}} & \textcolor{blue}{\bf{71.9\%}} & 12.2\% & 44.0\% & 7762 & 32547 & 442 & 823\\
    \cline{2-11}
    & \multirow{6}{*}{Online} & TC\_ODAL\cite{TC_ODAL} & 15.1\% & 70.5\% & 3.2\% & 55.8\% & 12970 & 38538 & 637 & 1716\\
    & & RMOT\cite{RMOT} & 18.6\% & 69.6\% & 5.3\% & 53.3\% & 12473 & 36835 & 684 & 1282\\
    & & oICF\cite{oICF} & 27.1\% & 70.0\% & 6.4\% & 48.7\% & 7594 & 36757 & 454 & 1660\\
    & & SCEA\cite{SCEA} & 29.1\% & 71.1\% & 8.9\% & 47.3\% & 6060 & 36912 & 604 & \textcolor{red}{\bf{1182}}\\
    & & MDP\cite{MDP} & 30.3\% & \textcolor{red}{\bf{71.3\%}} & \textcolor{red}{\bf{13.0\%}} & \textcolor{red}{\bf{38.4\%}} & 9717 & \textcolor{red}{\bf{32422}} & 680 & 1500\\
    & & STAM & \textcolor{red}{\bf{34.3\%}} & 70.5\% & 11.4\% & 43.4\% & \textcolor{red}{\bf{5154}} & 34848 & \textcolor{red}{\bf{348}} & 1463\\
    \hline
    \hline
    \multirow{10}{*}{MOT16} & \multirow{6}{*}{Offline} & JPDA\_m\cite{JPDA} & 26.2\% & 76.3\% & 4.1\% & 67.5\% & \textcolor{blue}{\bf{3689}} & 130549 & 365 & 638\\
    & & SMOT\cite{SMOT} & 29.7\% & 75.2\% & 5.3\% & 47.7\% & 17426 & 107552 & 3108 & 4483\\
    & & CEM\cite{CEM} & 33.2\% & 75.8\% & 7.8\% & 54.4\% & 6837 & 114322 & 642 & 731\\
    & & MHT\_DAM\cite{MHT} & 45.8\% & 76.3\% & 16.2\% & 43.2\% & 6412 & 91758 & 590 & 781\\
    & & JMC\cite{JMC} & 46.3\% & 75.7\% & 15.5\% & \textcolor{blue}{\bf{39.7\%}} & 6373 & 90914 & 657 & 1114\\
    & & NOMT\cite{NOMT} & \textcolor{blue}{\bf{46.4\%}} & \textcolor{blue}{\bf{76.6\%}} & \textcolor{blue}{\bf{18.3\%}} & 41.4\% & 9753 & \textcolor{blue}{\bf{87565}} & \textcolor{blue}{\bf{359}} & \textcolor{blue}{\bf{504}}\\
    \cline{2-11}
    & \multirow{4}{*}{Online} & OVBT\cite{OVBT} & 38.4\% & \textcolor{red}{\bf{75.4\%}} & 7.5\% & 47.3\% & 11517 & 99463 & 1321 & 2140\\
    & & EAMTT\cite{EAMTT} & 38.8\% & 75.1\% & 7.9\% & 49.1\% & 8114 & 102452 & 965 & 1657\\
    & & oICF\cite{oICF} & 43.2\% & 74.3\% & 11.3\% & 48.5\% & \textcolor{red}{\bf{6651}} & 96515 & \textcolor{red}{\bf{381}} & \textcolor{red}{\bf{1404}}\\
    & & STAM & \textcolor{red}{\bf{46.0\%}} & 74.9\% & \textcolor{red}{\bf{14.6\%}} & \textcolor{red}{\bf{43.6\%}} & 6895 & \textcolor{red}{\bf{91117}} & 473 & 1422\\
    \hline
  \end{tabular}
  \caption{Quantitative results of our method (denoted by STAM) and several state-of-the-art MOT trackers on MOT15 and MOT16 test sequences. Results are divided into two groups, i.e. online tracking and offline tracking. \textcolor{red}{\bf{red}} and \textcolor{blue}{\bf{blue}} values in blod highlight the best results of online and offline methods respectively.
  '$\uparrow$' means that higher is better and '$\downarrow$' represents that lower is better.}
  \label{table:result}
\end{table*}

\subsection{Comparisons with state-of-the-art methods}
We compare our algorithm, denoted by STAM, with several state-of-the-art MOT tracking methods on the test sequences of MOT15 and MOT16 benchmarks.
All the compared state-of-the-art methods and ours use the same public detections provided by the benchmark for fair comparison.
Table \ref{table:result} presents the quantitative comparison results \footnote{The quantitative tracking results of all these trackers are available at the website http://motchallenge.net/results/2D\_MOT\_2015/ and http://motchallenge.net/results/MOT16/.}.

{\bf MOT15 Results.} Overall, STAM achieves the best performance in MOTA and IDS among all the online and offline methods. In terms of MOTA, which is the most important metric for MOT, STAM improves 4\% compared with MDP, the best online tracking method that is peer-reviewed and published. Note that our method works in pure online mode and dose not need any training data with ground truth annotations. While MDP performs training with sequences in the similar scenario and its ground truth annotations for different test sequences. Besides, our method produce the lowest IDS among all methods, which demonstrates that our method can handle the interaction among targets well. Note that the CNNTCM and SiameseCNN also utilize CNNs to handle MOT problem but in offline mode. What's more, their methods requir abundant training data for learning siamese CNN. The better performance compared to these CNN-based offline methods provides strong support on the effectiveness of our online CNN-based algorithm.

{\bf MOT16 Results.} Similarly, STAM achieves the best performance in terms of MOTA, MT, ML, and FN among all online methods. Besides, the performance of our algorithm in terms of MOTA is also on par with state-of-the-art offline methods.

On the other hand, our method produces slightly more Frag than some offline methods, which is a common defect of online MOT methods due to long term occlusions and severe camera motion fluctuation.

\section{Conclusion}
In this paper, we have proposed a dynamic CNN-based online MOT algorithm that efficiently utilizes the merits of single object trackers using shared CNN features and ROI-Pooling. In addition, to alleviate the problem of drift caused by frequent occlusions and interactions among targets, the spatial-temporal attention mechanism is introduced. Besides, a simple motion model is integrated into the algorithm to utilize the motion information. Experimental results on challenging MOT benchmarks demonstrate the effectiveness of the proposed online MOT algorithm.

{\bf Acknowledgement:} This work is supported by the National Natural Science Foundation of China (No.61371192), the Key Laboratory Foundation of the Chinese Academy of Sciences (CXJJ-17S044), the Fundamental Research Funds for the Central Universities (WK2100330002), SenseTime Group Limited, the General Research Fund sponsored by the Research Grants Council of Hong Kong (Project Nos. CUHK14213616, CUHK14206114, CUHK14205615, CUHK419412, CUHK14203015, CUHK14207814, and CUHK14239816), the Hong Kong Innovation and Technology Support Programme (No.ITS/121/15FX), and ONR N00014-15-1-2356.



{\small
\bibliographystyle{ieee}
\bibliography{reference_abrv}
}

\end{document}